# Information-Theoretical Learning of Discriminative Clusters for Unsupervised Domain Adaptation

**Yuan Shi**  YUANSHI@USC.EDU
U. of Southern California, Los Angeles, CA 90089 USA

**Fei Sha**  FEISHA@USC.EDU
U. of Southern California, Los Angeles, CA 90089 USA

## Abstract

We study the problem of unsupervised domain adaptation, which aims to adapt classifiers trained on a labeled source domain to an unlabeled target domain. Many existing approaches first learn domain-invariant features and then construct classifiers with them. We propose a novel approach that jointly learn the both. Specifically, while the method identifies a feature space where data in the source and the target domains are similarly distributed, it also learns the feature space *discriminatively*, optimizing an information-theoretic metric as an proxy to the expected misclassification error on the target domain. We show how this optimization can be effectively carried out with simple gradient-based methods and how hyperparameters can be cross-validated without demanding any labeled data from the target domain. Empirical studies on benchmark tasks of object recognition and sentiment analysis validated our modeling assumptions and demonstrated significant improvement of our method over competing ones in classification accuracies.

## 1. Introduction

Supervised learning algorithms often assume that the training and the test data are randomly sampled from the same joint distribution. While the assumption facilitates rigorous theoretical analysis and empirical comparison of different algorithms, its validity is often challenged outside of laboratory settings. In real-world applications, there are many factors causing a mismatch between the training and the test data. For instance, imagine developing a face detection system for Facebook mobile users. A tuned classifier on images captured by webcams could be applied to images from mobile phones. In this case, the imaging conditions vary significantly due to background illumination, motion blurring, pose, etc.

Techniques for addressing learning problems with mismatched distributions are often referred as domain adaptation, or sometimes transfer learning (Daumé III & Marcu, 2006; Pan & Yang, 2010; Quiñonero-Candela et al., 2009). The *source* domain refers to the labeled training data, while the *target* domain refers to the test data. When there is no labeled data from the target domain to help learning classifiers, the problem setting is termed *unsupervised domain adaptation*.

Unsupervised domain adaptation is especially challenging as the target domain does not provide explicitly any information on how to optimize classifiers. Note that the objective of domain adaptation is to derive a classifier for the unlabeled (target) data from the labeled (source) data. This goal sets domain adaptation apart from semi-supervised learning, whose primary goal is to improve the performance on the labeled data with unlabeled data (Chapelle et al., 2006). The difference is subtle yet fundamental. For example, model selection or cross-validation using classification accuracy on the target domain is generally impossible.

Existing approaches thus rely on making strong assumptions on how the data distribution have shifted between the two domains in order to derive classification rules for the target domain. For instance, in covariate shift (Shimodaira, 2000; Bickel et al., 2007; Huang et al., 2007), the marginal distributions of the features are different across domains while the posterior distribution of the label remains the same. This naturally leads to a two-stage learning paradigm:

the labeled instances from the source domain are first weighted so as to compensate the difference in marginal distributions. Then, a classifier is trained using the labels and then applied to the unlabeled data.

Other works have also followed similar paradigms (Pan et al., 2011; Gopalan et al., 2011). In the structural correspondence learning, the original features are first augmented with features that are more likely to be domain invariant and then a classifier is trained (Blitzer et al., 2006). The augmenting features are linear transformation of the original features. Alternatively, in deep learning architecture for domain adaptation, the augmenting features are highly nonlinear transformation of the original ones (Glorot et al., 2011).

Underlying all these methods is the assumption that there exists a domain-invariant feature space such that the marginal distributions of two domains are the same in the new feature space. Thus, classifiers learnt in the new space will perform equally well on both the source and the target. Theoretical analysis have showed that the loss on the target domain for any labeling functions depends on the difference between the marginal distributions, thus justifying the need to identify a feature space such that the two domains look alike to each other (Ben-David et al., 2007; Mansour et al., 2009).

We hypothesize that this view and practice of two-stage learning are restrictive. One possible fallacy is that maximizing the similarity in marginal distributions bear no direct consequence on (dis)similarities between posterior distributions. Thus, if there are multiple feature spaces where the source and the target domains have similar marginals, there is no reason to believe that a classifier trained on an arbitrarily chosen one would necessarily perform well on the target domain. As an extreme case, projecting features into irrelevant feature dimensions would make the two domains look very much alike!

Hence, the caveat is to retain *discriminative* information for constructing classifiers while we search for the domain-invariant feature space. This seems relatively straightforward to achieve if all we care is the discriminative information about the labels in the source domain. However, our main goal is to have good classifiers for the target domain. Thus, our challenge is **how to be discriminative without labels?**

To address this challenge, we propose a novel learning algorithm for unsupervised domain adaptation. As opposed to existing two-stage approaches where new feature spaces and classifiers are separately optimized, our approach combines the two in a single stage. Moreover, the new feature space is discriminative with respect to the target domain. We give a brief account in the following, leaving details to sections 2 and 3.

**Main Idea** We assume *discriminative clustering*, namely, data in both the source and the target domains are tightly clustered and clusters corresponds class boundaries. For the same class, the clusters from the two domains are geometrically close to each other. Leveraging these assumptions, our formulation of learning the optimal feature space balances two forces: maximizing domain similarity that makes the source and the target domains look alike, and (approximately) minimizing the expected classification error on the target domain. We define those two forces with information-theoretical quantities: the domain similarity being the negated mutual information between all data and their binary domain labels (SOURCE versus TARGET) and the expected classification error being the negated mutual information between the target data and its clusters (ie class) labels estimated from the source data. These two quantities are directly motivated by the nearest neighbor classifiers we use in the new feature space.

We show how simple gradient-based methods can be effectively used for numerical optimization to learn the optimal feature space. We evaluated extensively our approach on two benchmark tasks: visual object recognition and sentiment analysis of product reviews. On both of them, the proposed approach outperforms other state-of-the-arts methods significantly.

**Contributions** To summarize, we contribute to domain adaptation by advocating discriminative clustering as a possible mechanism for adaptation; cf. section 2. We hypothesize that existing approaches of two-stage learning can be significantly improved by taking those cluster structures into consideration. Thus, we propose an one-stage approach *jointly* learning a domain-invariant feature space and optimizing information-theoretic metrics directly related to discriminative classification on the target domain; cf. section 3. Our empirical results support strongly our modeling assumptions and hypothesis; cf. section 4.

## 2. Discriminative Clustering for Domain Adaptation

At the core of our approach is the assumption of discriminative clustering. Specifically, we assume that, in a suitable feature space, 1) SEPARATION. Data in the source and the target domains are *discriminatively* clustered, where the cluster ids correspond to class labels; 2) ALIGNMENT. The clusters from the two domains that correspond to the same label are geomet-rically close. Fig. 1 illustrates these two assumptions and how they can be exploited for adaptation.

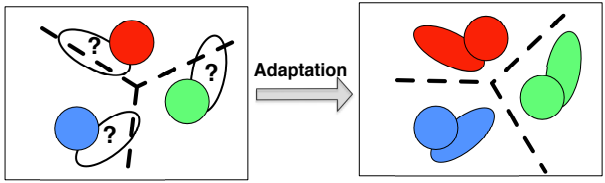

*Figure 1.* Schematic illustration of our main idea on exploiting *discriminative clustering* for unsupervised domain adaptation, cf. section 2. Data in the source domain (within circle shapes) and the target domain (within oval shapes) are tightly clustered, corresponding to their class boundaries. Moreover, clusters from the two domains are "aligned" if they correspond to the same class. Assuming and exploiting such structures in the data, classifier boundaries for the source domain (dashed lines in the left diagram) are adapted *discriminatively* to the target domain (dashed lines in the right diagram), minimizing the expected classification errors on the target domain. The target data is then classified with adapted classifiers. See section 3 for details on how the errors can be approximated using information-theoretical quantities such as mutual information, without using labels.

Arguably, the assumptions are more "relaxed" than those in existing works for adaptation. Specifically, they do not imply that the marginal distributions are the same across domains and certainly do not imply the same posterior distributions either. In fact, these assumptions are readily satisfiable in applications.

For example, many datasets exhibit multi-modal marginal distributions where the modes correspond to class labels, particularly if these data are sampled from a generative process of mixture models.

As we will show in the following, these two assumptions allow us to define quantitively what to be optimized – in our case, we would like to identify a domain-invariant feature space such that the expected misclassification error on the target data is minimized. Despite the paucity in labels from the target domain, we will show how the ALIGNMENT assumption will allow us to define a proxy to the error so as to be optimized.

## 3. Proposed Approach

In what follows, we are given $\mathsf{N}$ labeled instances from the source domain: $\{(\boldsymbol{x}_s, y_s)\}$ where $\boldsymbol{x}_s \in \mathcal{X} \subset \mathbb{R}^{\mathsf{D}}$ and $y_s$ takes a value from $\mathsf{C}$ class labels: $y_s \in \mathcal{Y} = \{1, 2, \ldots, \mathsf{C}\}$. We also have $\mathsf{M}$ unlabeled instances from the target domain: $\{\boldsymbol{x}_t\}$ where $\boldsymbol{x}_t \in \mathcal{X}$. For simplicity, we assume $\boldsymbol{x}_t$ and $\boldsymbol{x}_s$ have the same domain $\mathcal{X}$, thus the same dimensionality. Extensions to more general cases are possible, analogous to (Kulis et al., 2011).

Our objective is to construct a classifier $f : \boldsymbol{x} \in \mathcal{X} \to y \in \mathcal{Y}$. We would like the classifier performs well on the target domain $\mathcal{D}_T$ from which $\boldsymbol{x}_t$ is sampled. This is inherently an ill-posed problem as we do not have any labels from the target domain.

To overcome this difficulty, we leverage the *discriminative clustering* assumptions which we have previously described. We assume that there is a latent feature space $\boldsymbol{z} \in \mathbb{R}^{\mathsf{d}}$ such that i) data in the source and target domains form well-separated clusters and the clusters correspond to labels; ii) the clusters from the source domain are geometrically close to those from the target domain if they are from the same labels.

We show how these assumptions can be used to derive information theoretical quantities which reflect data characteristics in each domain. These quantities are parameterized in terms of the latent feature which is in turn a linear transformation of the original feature $\boldsymbol{x}$. We then show how to combine these quantities so that the optimal linear transformation can be learnt from data. We begin by describing a few key notions.

### 3.1. Conditional models in the feature space

We consider the latent feature space induced by a linear transformation $\boldsymbol{L} \in \mathbb{R}^{\mathsf{d} \times \mathsf{D}}$. In the new feature space, we use k-nearest neighbors (kNN) to classify as we have assumed that data form well-separated clusters. Moreover, we choose $k = 1$ to avoid cross-validating this parameter.

The squared distance between two points $\boldsymbol{x}_i$ and $\boldsymbol{x}_j$ in this feature space is thus given by

$$d_{ij}^2 = \|\boldsymbol{L}\boldsymbol{x}_i - \boldsymbol{L}\boldsymbol{x}_j\|_2^2 = (\boldsymbol{x}_i - \boldsymbol{x}_j)^{\mathrm{T}} \boldsymbol{M} (\boldsymbol{x}_i - \boldsymbol{x}_j) \quad (1)$$

where $\boldsymbol{M} = \boldsymbol{L}^{\mathrm{T}} \boldsymbol{L}$ defines a (low-rank) Mahalanobis distance metric in the original space.

Given a point $\boldsymbol{x}_i$ and a set of data points $\{\boldsymbol{x}_j\}$, we use the following model

$$p_{ij} = \frac{e^{-d_{ij}^2}}{\sum_{j \neq i} e^{-d_{ij}^2}} \quad (2)$$

to define the conditional probability of having $\boldsymbol{x}_j$ as $\boldsymbol{x}_i$'s nearest neighbor.

The above conditional model has been used in many contexts, including metric learning (Goldberger et al., 2004), dimensionality reduction (Hinton & Roweis, 2002), etc. Characterizing how close a point $\boldsymbol{x}_i$ is to other points, this model gives rise to an estimate of the posterior $p(y_i = c|\boldsymbol{x}_i)$ for labeling $\boldsymbol{x}_i$ with the class label $c$, assuming the class labels of $\{\boldsymbol{x}_j\}$ are known,

$$\hat{p}_{ic} = \sum_{j \neq i} p_{ij} \delta_{jc} \qquad (3)$$

where $\delta_{jc}$ is 1 if $\boldsymbol{x}_j$'s label is $c$, and 0 otherwise. Since $p_{ij}$ is a normalized probability, $\hat{p}_{ic}$ is normalized too. For example, if the label of $\boldsymbol{x}_i$ is known, $\sum_c \hat{p}_{ic} \delta_{ic}$ would be the probability of correctly classifying $\boldsymbol{x}_i$.

### 3.2. Discriminative clustering in the source

To derive a classifier that can perform well on the target domain, we would certainly need the classifier to perform well on the source domain because we have assumed that the two domains share similar clustering structures. Thus, our first desideratum is to minimize the expected classification error on the source domain, when we classify it using 1-NN. This error is estimated using the empirical average of the leave-one-out accuracy for any given point $\boldsymbol{x}_s$ in the source domain $\mathcal{D}_S$:

$$\varepsilon_s = 1 - \frac{1}{\mathsf{N}} \sum_s \sum_c \hat{p}_{sc} \delta_{sc} \qquad (4)$$

Note that, if we minimize this error only and ignore the target domain, we will arrive at the metric learning technique in (Goldberger et al., 2004).

### 3.3. Discriminative clustering in the target

Since we do not have labels on the target domain, we cannot define the expected classification error as we did in eq. (4) for the source domain. **How to be discriminative without using labels?**

Consider an instance $\boldsymbol{x}_t$ from the target domain and all the instances $\{\boldsymbol{x}_s\}$ from the source domain, the conditional model $p_{ts}$ of eq. (2) gives rise to the probability of having a particular $\boldsymbol{x}_s$ as the nearest neighbor of $\boldsymbol{x}_t$. Using this conditional model as well as the source labels to compute the posterior as in eq. (3) would not be the correct posterior for the target domain. However, if our assumptions about two sets of clusters being geometrically close indeed hold in the dataset, then the estimation $\hat{p}_{tc}$ should be close to the true posterior.

If $\hat{p}_{tc}$ approximates the true posterior well and our assumption that the target data is well clustered, then we can reasonably expect that the C-dimensional probability vector $\hat{\boldsymbol{p}}_t = [\hat{p}_{t1}, \hat{p}_{t2}, \ldots, \hat{p}_{t\mathsf{C}}]$ should look like an *ideal* posterior probability vector $[0, 0, \ldots, 1, \ldots, 0]$ where the only nonzero element 1 occurs at the position corresponding to the correct label.

Since we do not know the true label, we cannot measure directly the similarity of $\hat{\boldsymbol{p}}_t$ to the correct and ideal posterior vector. Nonetheless, we can express our desideratum as reducing the entropy of $\hat{\boldsymbol{p}}_t$ such that it contains the least amount of confusing labels.

Let $H[\boldsymbol{p}]$ denote the entropy of a probability vector $\boldsymbol{p}$. If we minimize $\sum_t H[\hat{\boldsymbol{p}}_t]$ only, we could arrive at a degenerate solution where every point $\boldsymbol{x}_t$ is assigned to the same class. To avoid this, we instead maximize the mutual information between the data and the estimated label $\hat{Y}$ using $\hat{\boldsymbol{p}}$,

$$I_t(X; \hat{Y}) = H[\hat{\boldsymbol{p}}_0] - \frac{1}{\mathsf{M}} \sum_t H[\hat{\boldsymbol{p}}_t] \qquad (5)$$

and the prior distribution $\hat{\boldsymbol{p}}_0$ is given by $\hat{\boldsymbol{p}}_0 = 1/\mathsf{M} \sum_t \hat{\boldsymbol{p}}_t$. Note that using the empirical distribution of the labels in the source domain to estimate the prior $\hat{\boldsymbol{p}}_0$ could still lead to degenerate solutions when the labels are uniformly distributed.

Minimizing the entropy (or similarly, maximizing the mutual information) has been previously studied in the context of (discriminative) clustering, cf. (Gomes et al., 2010; Dhillon et al., 2003). This criterion will identify a feature representation that classifiers can use to achieve a low lower-bound of misclassification error, due to Fano's inequality (Fisher III & Principe, 1998).

### 3.4. Discriminability: source versus target

The previous discussion on discriminative clustering in the target domain hinges on the assumption that clusters for the source and the target domain should not be too far from each other. We quantify this notion more precisely in the following. Conceptually, this notion is similar to the idea in existing works to make marginal distributions similar across domains.

Why such notion is desirable? To use the source domain's labels as an proxy to estimate the posterior probabilities for the target data (as in eq. (3)), we would desire the source and the target domain share some common probability supports in the feature space. In particular, consider the case we classify two instances $\boldsymbol{x}_t$ and $\boldsymbol{x}_{t'}$ from the target domain. They are deemed to have the same label $c$ if there are plenty of labeled source data in class $c$ in their neighborhoods. Then we would expect that with high likelihood, $\boldsymbol{x}_t$ and $\boldsymbol{x}_{t'}$ are in each other's nearest neighbors too — otherwise, the cluster corresponding to class $c$ in the target domain would not be very "tight".

Having instances from both domains in $\boldsymbol{x}_t$'s nearest neighborhoods thus entails the following. If we create a binary classification problem and assign $q_i = 1$ if $\boldsymbol{x}_i$ is from the source and $q_i = 0$ if $\boldsymbol{x}_i$ from the target, then given $\boldsymbol{x}_i$, we cannot determine well above chance level where this instance comes from.

Instead of constructing an actual binary classifier, we express our desideratum as minimizing the mutual information between the data instance $X$ and its (binary) domain label $Q$. Analogous to eq. (5), the mutual information is given by,

$$I_{st}(X;Q) = H[\hat{\boldsymbol{q}}_0] - \frac{1}{\mathsf{N}+\mathsf{M}} \sum_i H[\hat{\boldsymbol{q}}_i] \qquad (6)$$

where $\hat{\boldsymbol{q}}_i$ is the two-dimensional posterior probability vector of assigning $\boldsymbol{x}_i$ to either the source or the target, given *all* other data points from the two domains. Concretely, the probability is computed according to eq. (3), except the class label $\delta_{jc}$ being replaced by the domain label of $\boldsymbol{x}_j$. The estimated prior distribution $\hat{\boldsymbol{q}}_0$ is computed as $1/(\mathsf{N}+\mathsf{M})\sum_i \hat{\boldsymbol{q}}_i$.

One might wonder why we do not compute and minimize the expected error as in the source domain classification eq. (4). This is because we would like to leave some room for the possibility that a certain portion of data in either domain could be "outliers" to the other domain, and thus indeed distinguishable with respect to their origins. Minimizing domain classification error would have the adverse effect of forcing the two domains to be exactly the same. For instance, a degenerate solution would be to map every point to the origin of the feature space.

We mention in passing that it is found that the accuracy of a binary domain classifier reflects similarities between domains (Blitzer et al., 2007), thus approximating the original intractable combinatorial measure of similarities (Ben-David et al., 2007).

### 3.5. Learning and model selection

We have described three information-theoretical quantities: classification accuracies on the source domain $\varepsilon_S$ of eq. (4), discriminative clustering on the target $I_t(X;\hat{Y})$ of eq. (5), and discriminability between the source and the target $I_{st}(X;Q)$ of eq. (6).

These quantities have been derived from our assumptions about the source and target domains, specifically, the discriminative clustering structures. They are all parameterized in the linear transformation $\boldsymbol{L}$.

We learn the optimal $\boldsymbol{L}$ by balancing these quantities with the following optimization problem

$$\begin{aligned}\text{minimize} &\quad -I_t(X;\hat{Y}) + \lambda I_{st}(X;Q) \\ \text{subject to} &\quad \text{Trace}(\boldsymbol{L}^{\mathsf{T}}\boldsymbol{L}) \leq \mathsf{d}\end{aligned} \qquad (7)$$

where the constraint is to control the scale of distances computed using $\boldsymbol{L}$.

The regularization coefficient $\lambda$ needs to be cross-validated. We choose the optimal $\lambda$ that attains the minimum of $\varepsilon_S$. Intuitively, $\varepsilon_S$ is defined on the source domain with labeled data and thus, more sensible to be used for model selection (Other ways of combining these quantities were also experimented, though the above performs the best in practice.)

We comment briefly on the difference between our formulation and the entropy minimization framework for semi-supervised learning (Grandvalet & Bengio, 2005). Their goal is to reduce uncertainty of labeling the unlabeled data. Thus, they use only the entropy term eq. (3). More distinctively, they do not need to make the two domains look alike thus there is no need for them to learn a feature space, nor to include a term to minimize the discriminability between the domains.

### 3.6. Numerical Optimization

Eq. (7) is non-convex optimization. We use gradient-based methods to optimize the objective function. While in theory the methods are susceptible to local optimum, we use heuristics to initialize: either the PCA of the target domain data, or the low-rank factorization of a discriminatively trained metric on the source data, such as the one in large margin nearest neighbor (LMNN) (Weinberger & Saul, 2009). In most cases, these heuristics work well and lead to substantially improved results over initialization points. Details are described in the Supplementary Material.

### 3.7. Extensions

When the target domain has a few labeled instance, the domain adaptation problem is referred as *semi-supervised adaptation*. Our approach can be readily extended to incorporate those labeled target domain instances. Details, including experimental results are described in the Supplementary Material.

## 4. Experimental Results

We evaluate the proposed method on two benchmark tasks: object recognition and sentiment analysis of product reviews. We compare the method to baselines and other recently proposed ones for unsupervised domain adaptation (Gopalan et al., 2011; Blitzer et al., 2006; Pan et al., 2011). In the Supplementary Material, we report results on semi-supervised adaptation, where the target domain has a few labeled instances.

### 4.1. Setup

We start by describing the datasets for the two tasks.

**Object recognition**. We use four databases of object images: Caltech-256 (Griffin et al., 2007), Amazon (images from online merchants's catalogues), Webcam (low-resolution images by web cameras), and DSLR (high-resolution images by digital SLR cameras). The last three datasets were studied in (Gopalan et al., 2011; Saenko et al., 2010). Caltech-256 is added to increase the diversity of the domains.

We treat each dataset as a domain. There are 10 common object categories: backpack, coffee-mug, calculator, computer-keyboard, computer-monitor, computer-mouse, head-phones, laptop-101, touring-bike, and video-projector. There are 2533 images in total, with 8 to 151 images per category per domain.

Following the experimental protocols in previous work (Saenko et al., 2010), we extract SURF features (Bay et al., 2006) and encode each image with a 800-bin histogram (the codebook is trained from a subset of Amazon images). The histograms are first normalized to have zero mean and unit standard deviation in each dimension.

For each pair of source and target domains, we conduct experiments in 20 random trials. In each trial, we randomly sample labeled data in the source domain as the training set, and unlabeled data in the target domain as the testing set. For semi-supervised domain adaptation, we also sample a few labeled examples in the target domain to augment the training set, see the Supplementary Material for details.

**Sentiment analysis**. We use the dataset that consists of Amazon product reviews on four product types: kitchen appliances, DVDs, books and electronics (Blitzer et al., 2007). Each product type is used as a separate domain. Each domain has 1,000 positive and 1,000 negative reviews. To reduce computational cost, we select top 400 words of the largest mutual information with the labels. We then represent each review with a 400-dimensional vector of term counts (ie, bag-of-words). The vectors are normalized to have zero mean and unit standard deviation in each dimension.

For each pair of source and target domains, we conduct experiments in 10 random trials. In each trial, we randomly sample 1,600 labeled data in the source domain as the training set, and all data in the target domain as the testing set.

**Classification** We learn the feature transformation $L$ by solving the optimization problem eq. (7). We then transform all the data using the matrix and apply 1-nearest neighbor (1-NN) to classify instances from the target domain. 1-NN is used to avoid tuning the number of nearest neighbors. (In the Supplementary Material, we also report results of using SVMs.)

**Hyperparameter tuning** Our method has two hyper-parameters: the dimensionality of the new feature subspace and the regularization coefficient $\lambda$ in eq. (7). We cross-validate them using the model selection procedure described in section 3.5. The range of search for the dimensionality is $\{20, 40, 70, 100\}$ and $\{0, 0.25, 1, 4, 16, 64\}$ for $\lambda$.

For baselines and other methods we have compared to, if there are hyper-parameters to be tuned, we either follow the procedures in those algorithms or give those methods the benefits of doubts by reporting their ***best*** performance by using labels from the target domain.

### 4.2. Results on unsupervised adaptation

We compare extensively to several methods.

- Baselines. We compare to **PCA**, where we project all data into the PCA directions computed on the *target domain*. We also compare to **LMNN** (Weinberger & Saul, 2009), where we train a large margin nearest neighbor classifier using only the *source* labeled data. Neither of these methods is developed for domain adaptation and their performances on target domains are indeed inferior to other methods, and especially ours.
- Transfer Component Analysis (**TCA**) (Pan et al., 2011). This method finds a low-dimensional linear projection such that the source and the target domains have similar marginal distributions, regularized by preserving variances in all the data. To measure similarities in marginals, the method maps data to a kernel feature space. We use Gaussian RBF kernels.
- Structural Correspondence Learning (**SCL**) (Blitzer et al., 2006). This method augments original features with linearly transformed features. The linear transformation is computed as the principal directions of parameters in binary classifiers predicting whether pivot features are present or not. In our experiments, we have used all 400 features as pivot features. We then train SVMs with the augmented feature vectors on the source domains and apply the resulting classifiers to the target domains.
- Geodesic Flow Subspaces (**GFS**) (Gopalan et al., 2011). This method interpolates (on Grassman manifold) between the PCA subspaces computed on the source and the target domains respectively. The interpolated subspaces are then used to transform the original features to form super-vectors. The dimensionality of the super-vectors is then

Table 1. Classification accuracies on target domains with *unsupervised* adaptation

| Domains | PCA | TCA | GFS | LMNN | Metric | Ours |
|---|---|---|---|---|---|---|
| DSLR → Webcam | 80.6±0.5 | 66.2±0.5 | 75.5±0.4 | 81.3±0.4 | 55.6±0.7 | **83.6±0.5** |
| DSLR → Amazon | 35.1±0.3 | 31.4±0.2 | 35.7±0.5 | **42.3±0.3** | 30.3±0.8 | 39.6±0.4 |
| Caltech → DSLR | 36.6±1.2 | 33.1±0.8 | 36.5±0.9 | 37.2±1.1 | 35.0±1.1 | **44.4±1.2** |
| Caltech → Amazon | 37.7±0.5 | 34.9±0.4 | 37.9±0.5 | 43.2±0.4 | 33.7±0.8 | **49.2±0.6** |
| Amazon → Webcam | 33.1±0.6 | 26.5±0.8 | 32.8±0.7 | 35.2±0.8 | 36.0±1.0 | **38.5±1.3** |
| Amazon → Caltech | 35.9±0.3 | 29.3±0.3 | 36.1±0.5 | 37.6±0.4 | 27.3±0.7 | **40.0±0.4** |

reduced before applying 1-NN for classification.

- Metric Learning (**Metric**) (Saenko et al., 2010). This method learns a metric measuring the distance between data points using the correspondence information between the source and the target domains. Specifically, the correspondence is defined as data points with the same labels. Thus, this method uses *labels* from the target domains. Despite that, our results will show our method still outperforms **Metric**.

Table 1 and Table 2 summarize the classification accuracies as well as standard errors of all the above methods, as well as ours (we did not apply **SCL** to object recognition as it is difficult to define what pivot features are for those types of data). We had chosen a subset of all pairs for saving experiment time. The best performing algorithm(s) (statistical significant up to one standard error) for each pair are in bold font.

In Table 1 on object recognition, our method performs the best on 5 out of 6 pairs, outperforming other competing methods with a large margin. On the DSLR-Amazon pair, our method performs worse than **LMNN**, but still significantly better than others.

Of particular interest is that **LMNN** outperforms other methods specifically designed for domain adaptation (excluding ours). This confirms our hypothesis: the two-stage learning schemes adopted by **TCA** and **GFS** suffer from the fallacy that maximizing marginal similarity does not necessarily lead to well-performing classifiers on the target domain. In particular, we believe such methods could actually destroy discriminative information by forcing the domains to be similar.

The results thus support our argument that one-stage learning, namely identifying jointly discriminative clustering and low-dimensional feature spaces, is crucial for domain adaptation.

The results on sentiment analysis in Table 2 also strongly support similar conclusions. Note that both **SCL** and our methods outperform other methods significantly. Our methods perform better on 2 out of 4 pairs, thought slightly worse than **SCL** on the other two. Exploring strengths and weakness of each of these two methods is a subject of future research.

## 5. Related Work

Information-theoretical approach has been applied to semi-supervised learning (Grandvalet & Bengio, 2005) where the core idea is to reduce the confusability (among possible labels) on unlabeled data by classifiers trained on the labeled data. However, they have assumed that the data are drawn from the same distribution so there is no need to learn a domain-invariant feature space.

Rastrow et al. described an information-theoretical based criterion for model selection in domain adaptation (Rastrow et al., 2010). Model selection is a challenging problem when cross-validation is not possible due to the lack of labeled data on the target domain. However, their approach is two-stage: they refine model parameters on unlabeled data by minimizing the conditional entropy of the labeling function from the initial model tuned on the labeled source data. Consequently, their formulation does not learn an invariant feature space.

Our work is also related to the recent study of regularized information maximization for discriminative clustering (Gomes et al., 2010). The authors there used a parametric model to compute the posterior probabilities of assigning a data point to various clusters. The objective is to find clustering assignments of all data points such that the mutual information between the data and the cluster ids are maximized. While their work is generalized to semi-supervised clustering, they do not consider domain adaptation, which has fundamentally different goals and constraints from semi-supervised learning, as pointed out previously. In particular, the above-mentioned work does not learn a new feature space.

## 6. Conclusion

We propose an one-stage approach that *jointly* learns a domain-invariant feature space and optimizes information-theoretic metrics directly related to dis-criminative classification on the target domain. Our empirical results support the validity of our modeling assumptions that data in both source and target domains are discriminatively clustered. We show that existing approaches where learning feature is decoupled from learning discriminative classifiers, can be significantly improved by taking the clustering structures into consideration. For future work, we plan to study discriminatively learning of nonlinear feature transformation for domain adaptation.

Table 2. Classification accuracies on target domains with *unsupervised* adaptation

| Domains | PCA | SCL | TCA | GFS | LMNN | Ours |
|---|---|---|---|---|---|---|
| Kitchen → DVD | 66.1±0.7 | 73.2±0.6 | 64.9±0.5 | 67.9±1.0 | 70.8±0.5 | **75.4±0.6** |
| DVD → Books | 66.4±0.4 | **79.2±0.4** | 64±0.7 | 70.8±0.6 | 71.7±0.6 | 78.4±0.5 |
| Books → Electronics | 63.6±0.9 | 75.6±0.6 | 62.7±0.7 | 67.2±1.0 | 69.2±0.6 | **79.2±0.9** |
| Electronics → Kitchen | 71.8±0.4 | **84.5±0.5** | 69.5±0.7 | 75.8±1.2 | 77.3±0.6 | 82.9±0.5 |

## Acknowlegements

This work was partially supported by DARPA D11AP00278, NSF IIS-1065243 and a USC Annenberg Fellowship (Y. Shi).